\pdfoutput=1

\documentclass[11pt]{article}

\usepackage[]{acl}

\usepackage{times}
\usepackage{latexsym}

\usepackage[T1]{fontenc}

\usepackage[utf8]{inputenc}

\usepackage{microtype}

\usepackage{graphicx}
\usepackage{subcaption}
\usepackage{booktabs}
\usepackage{multirow}
\usepackage{enumitem}
\usepackage{amsmath,amssymb}
\usepackage{color}
\usepackage{soul}
\usepackage{natbib}

\title{A Study of Syntactic Multi-Modality in Non-Autoregressive Machine Translation}

\author{
Kexun Zhang$^{1}$\thanks{\ \ This work was conducted at Microsoft Research Asia.}, \ 
Rui Wang$^{2}$\thanks{\ \ Corresponding author: Rui Wang, ruiwa@microsoft.com, and Xu Tan, xuta@microsoft.com.}, \ 
Xu Tan$^{2\dag}$, \ 
Junliang Guo$^{2}$, \
Yi Ren$^{1}$,\
Tao Qin$^{2}$,\
Tie-Yan Liu$^{2}$ \\
$^{1}$Zhejiang University, 
$^{2}$Microsoft Research Asia\\
\texttt{$^{1}$\{kexunz,rayeren\}@zju.edu.cn}\\ \texttt{$^{2}$\{ruiwa,xuta,junliangguo,taoqin,tyliu\}@microsoft.com}
}
    
\begin{document}
\maketitle
\begin{abstract}
It is difficult for non-autoregressive translation (NAT) models to capture the multi-modal distribution of target translations due to their conditional independence assumption, which is known as the ``multi-modality problem'', including the lexical multi-modality and the syntactic  multi-modality. While the first one has been well studied, the syntactic multi-modality brings severe challenge to the standard cross entropy (XE) loss in NAT and is under studied. In this paper, we conduct a systematic study on the syntactic multi-modality problem. Specifically, we decompose it into short- and long-range syntactic multi-modalities and evaluate several recent NAT algorithms with advanced loss functions on both carefully designed synthesized datasets and real datasets. We find that the Connectionist Temporal Classification (CTC) loss and the Order-Agnostic Cross Entropy (OAXE) loss can better handle short- and long-range syntactic multi-modalities respectively. Furthermore, we take the best of both and design a new loss function to better handle the complicated syntactic multi-modality in real-world datasets. To facilitate practical usage, we provide a guide to use different loss functions for different kinds of syntactic multi-modality.

\end{abstract}

\section{Introduction}

Traditional Neural Machine Translation~(NMT) models predict each target token conditioned on previous generated tokens in an autoregressive way~\cite{vaswani2017attention}, resulting in high latency in inference. Non-Autoregressive Translation (NAT) models generate all the target tokens in parallel~\cite{gu2018non}, significantly reducing inference latency. A disadvantage of NAT is that it suffers from the multi-modality problem~\cite{gu2018non} when a source sentence corresponds to multiple correct translations~\cite{ott2018analyzing}.

There are two types of multi-modalities: the lexical one and the syntactic one. The former one has been adequately studied ~\cite{gu2018non,zhou2019understanding,ding2020understanding}, while the latter one brings severe challenges to the widely used cross entropy (XE) loss in NAT. With standard XE loss, the generated tokens are required to be strictly aligned with ground truth tokens in the same positions, which fails to provide positive feedback for correctly predicted words on different positions as shown in Fig.~\ref{fig:CE}. Therefore, advanced loss functions are introduced to provide better feedback for NAT training: Connectionist Temporal Classification (CTC) loss~\cite{libovicky2018end} considers all possible monotonic alignments between a generated sequence and the ground truth; Aligned Cross-Entropy (AXE) loss~\cite{ghazvininejad2020aligned} selects the best monotonic alignment; and Order-Agnostic cross entropy (OAXE) loss~\cite{du2021order} calculates the XE loss with the best alignment based on maximum bipartite matching algorithm.

Even if with those advanced loss functions, we find they do not perform consistently across datasets and languages. In addition, diverse grammar rules in natural language~\cite{comrie1989language} implies the existence of different kinds of syntactic multi-modality. Inspired by \citet{odlin2008handbook,jing2015mean,haitao15probability,liu2010dependency}, we categorize the syntactic multi-modality into two sub types: the long-range and short-range ones. The long-range multi-modality is mainly caused by long-range word order diversity (e.g., an adverbial of place may appear at the beginning or the end of a sentence). The short-range multi-modality is mainly caused by short-range word order diversity (e.g., an adverb may appear either in front of or behind the corresponding verb) and optional words (e.g., in some languages, determiners and prepositions may be optional~\cite{ott2018analyzing}). Based on the above categorization of syntactic multi-modality, we further ask two research questions: (1) Which kinds of syntactic multi-modality do these loss functions excel at respectively? (2) How to better address this problem by taking advantage of different loss functions?


In this paper, we conduct a systematic study to answer these questions:
\begin{itemize}[leftmargin=*]
\item Since the short-range and long-range syntactic multi-modalities are usually entangled together in real-world datasets, we first design synthesized datasets to decouple them to better evaluate existing NAT algorithms (\textsection\ref{sec:toy}). We find that the CTC loss~\cite{libovicky2018end} can better handle the short-range syntactic multi-modality while the OAXE loss~\cite{du2021order} is good at the long-range one. Though carefully designed, the synthesized datasets are still different from the real-world datasets. Accordingly, we further conduct analyses on real-world datasets (\textsection\ref{sec:real}), which also show consistent findings with that in synthesized datasets.  
\item We design a new loss function that takes the best of both CTC and OAXE, and performs better to handle the short- and long-range syntactic multi-modalities simultaneously (\textsection\ref{sec:better_solve}), as verified by experiments on benchmark datasets including WMT14 EN-DE, WMT17 EN-FI, and WMT14 EN-RU. Moreover, we further provide a practical guide to use different loss functions for different kinds of syntactic multi-modality (\textsection\ref{sec:better_solve}). 

\end{itemize}

\section{Background}
\label{sec:background}



\begin{figure}[t]
\centering

\begin{subfigure}[t]{0.48\textwidth}
    \includegraphics[width=\textwidth]{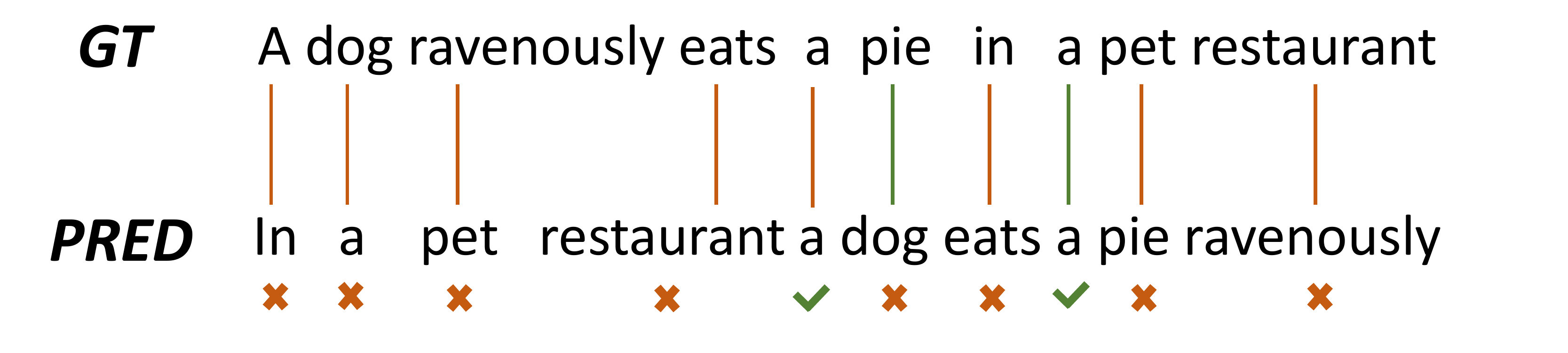}
    \caption{XE}
    \label{fig:CE}
\end{subfigure}

\begin{subfigure}[t]{0.48\textwidth}
    \includegraphics[width=\textwidth]{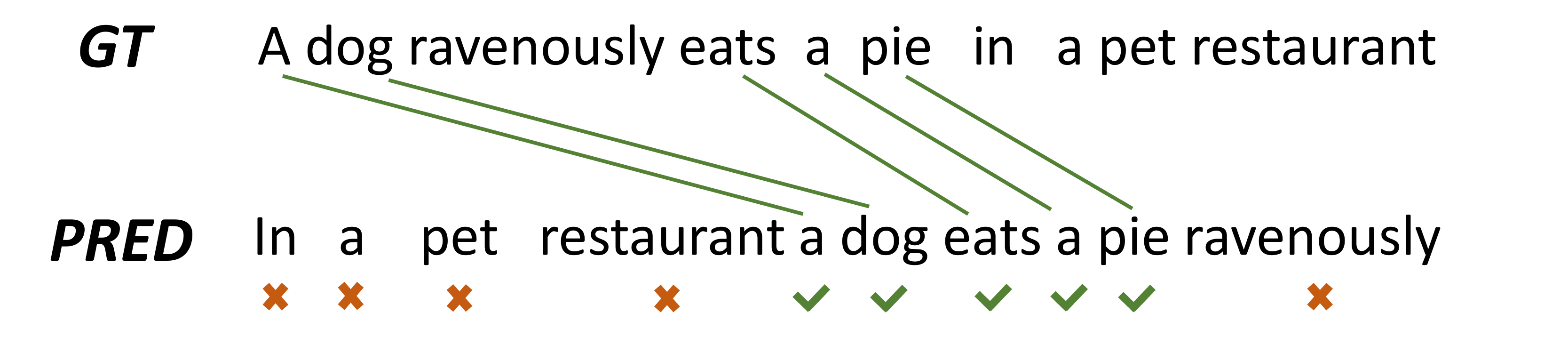}
    \caption{AXE}
    \label{fig:AXE}
\end{subfigure}

\begin{subfigure}[t]{0.48\textwidth}
    \includegraphics[width=\textwidth]{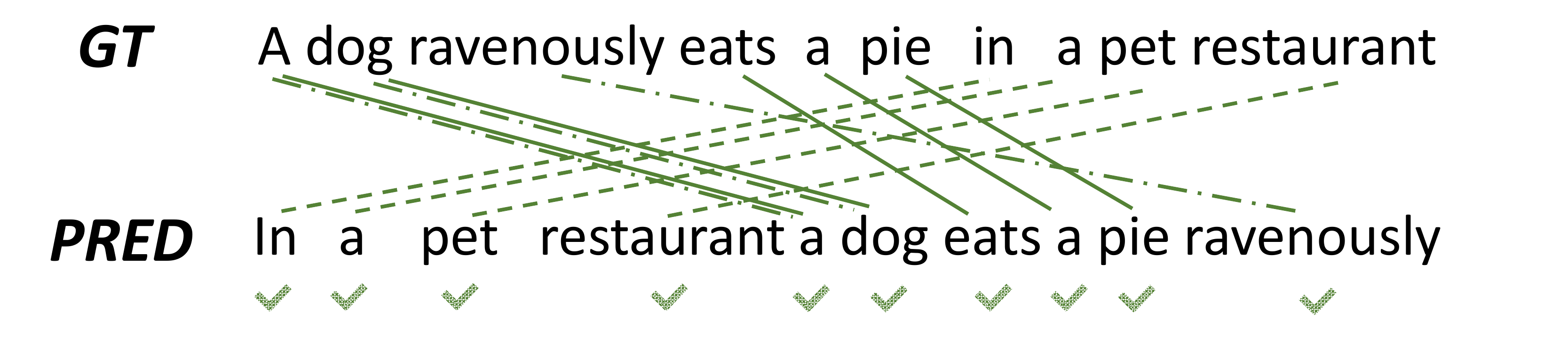}
    \caption{CTC, where solid, dash, and dot dash lines illustrate three possible alignments respectively.}
    \label{fig:CTC}
\end{subfigure}

\begin{subfigure}[t]{0.48\textwidth}
    \includegraphics[width=\textwidth]{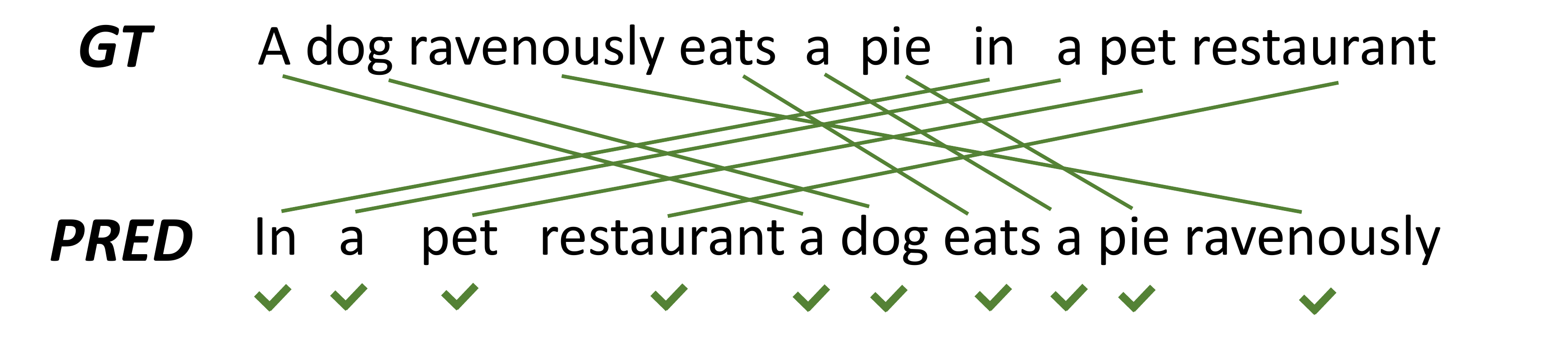}
    \caption{OAXE}
    \label{fig:OAXE}
\end{subfigure}
\caption{The illustration of different loss functions, where ``GT'' stands for ground truth, ``PRED'' stands for predicted sequence, the green check indicates that credit is provided to the token.}
\label{fig:LossFunction}
\vspace{-10pt}
\end{figure}

\paragraph{Non-Autoregressive Translation}
Given the source sentence $x=(x_1, x_2, ..., x_{T_{x}})$, traditional NMT model generates the target sentence $y=(y_1, y_2, ..., y_{T_{y}})$ from left to right and token by token: $P(y|x)=\prod_{t=1}^{T_{y}}P(y_{t}|y_{<t},x; \theta_{\textrm{enc}}, \theta_{\textrm{dec}})$,
where $y_{<t}$ indicates the target tokens generated before the $t$-th timestep, $T_{x}$ and $T_{y}$ denote the length of source and target sentence, $\theta_{\textrm{enc}}$ and $\theta_{\textrm{dec}}$ denote the encoder and decoder parameters respectively.
This autoregressive way suffers from high latency during inference. Non-Autoregressive Translation (NAT)~\cite{gu2018non} is proposed to reduce the inference time by generating the whole sequence in parallel,
$P(y|x)=P(T_{y}|x) \cdot \prod_{t=1}^{T_{y}}P(y_{t}|x; \theta_{\textrm{enc}}, \theta_{\textrm{dec}})$,
where $P(T_{y}|x)$ indicates the length prediction function. While the inference speed is boosted, the translation accuracy is sacrificed due to that target tokens are generated conditional independently.

\paragraph{Multi-Modality Problem} The multi-modality problem~\cite{gu2018non,zhou2019understanding} indicates that one source sentence may have multiple correct target translations, which brings challenges to NAT models as they generate each target token independently. Specifically, we categorize the multi-modality problem into two sub-problems, i.e., lexical and syntactic multi-modalities. 
The lexical multi-modality indicates that a source token can be translated into different target synonym tokens (i.e., ``thank you'' in English can be translated into both ``Danke'' or ``Vielen Dank'' in German), while the syntactic multi-modality indicates the inconsistency of word-orders (e.g., an adverb may appear either in front of or behind the corresponding verb) and the existence of optional words between source and target languages (e.g., in some languages, determiners and prepositions may be optional)~\cite{ott2018analyzing}. 
The lexical multi-modality problem has been adequately studied in recent works. Sequence-level knowledge distillation~\cite{gu2018non,zhou2019understanding} is shown capable to reduce the lexical diversity of the dataset and thus alleviate the problem. Some works also introduce extra loss functions such as KL-divergence~\citep{ding2020understanding} and bag-of-ngram~\citep{shao2020minimizing} to alleviate the lexical multi-modality problem.

On the contrary, 
there still lacks a systematic study on the syntactic multi-modality problem. Generally, it is difficult to solve this problem because the order and optional words vary across different languages. For example, the word order of Russian is quite flexible~\cite{kallestinova2007aspects}, thus the syntactic multi-modality may exist more frequently in Russian corpora. In contrast, the structure of English sentences is mostly subject–verb–object (SVO)~\cite{givon1983topic}, which results in less variation on word order.
In this paper, we categorize the syntactic multi-modality problem into short-range and long-range instances, and provide detailed analyses accordingly.

\paragraph{Loss Functions in NAT}
Standard cross-entropy~(XE) loss requires the predicted tokens to be strictly aligned with ground truth tokens, which fails to deal with the syntactic multi-modality problem. Different loss functions are proposed to solve the problem, and here we consider some most recent works. 
The CTC loss sums XE losses of all possible monotonic alignments and
has been widely used in speech recognition~\cite{graves2006connectionist,graves2013hybrid}, and the effectiveness of the CTC loss in NAT has been validated~\citep{libovicky2018end,gu2020fully}.
AXE~\cite{ghazvininejad2020aligned} selects the monotonic alignment between the predicted sequence and the ground truth with the minimum XE loss. 
OAXE~\cite{du2021order} further relaxes the position constraint and only considers the best alignment. The illustration for each loss function is provided in Fig.~\ref{fig:LossFunction}. Though effective in different datasets, these works ignore fine-grain features of the multi-modality problem such as short/long syntactic multi-modalities.
In this work, we analyse the performance of these loss functions in different syntactic scenarios, and provide a practical guide to use appropriate loss functions for different kinds of syntactic multi-modality.

\section{Analyses on Synthesized Datasets}
\label{sec:toy}

To make fine-grained analyses on the syntactic multi-modality problem, we first categorize it into long-range and short-range types, where the long-range one is mainly caused by long-range word order diversity, and the short-range one is mainly caused by short-range word order diversity and optional words. Then, we would like to evaluate the accuracy of different losses on different types of syntactic multi-modality. However, in real-world corpora, the different types are usually entangled, making it difficult to control and analyse one aspect without changing the other. 
Thus, we construct synthesized datasets based on phrase structure rules~\cite{chomsky1959certain} to manually control the degree of syntactic multi-modality in different aspects, and evaluate the performance of different existing techniques.

\subsection{Synthesized Datasets}

\begin{figure}[t]
\centering
\includegraphics[width=0.48\textwidth]{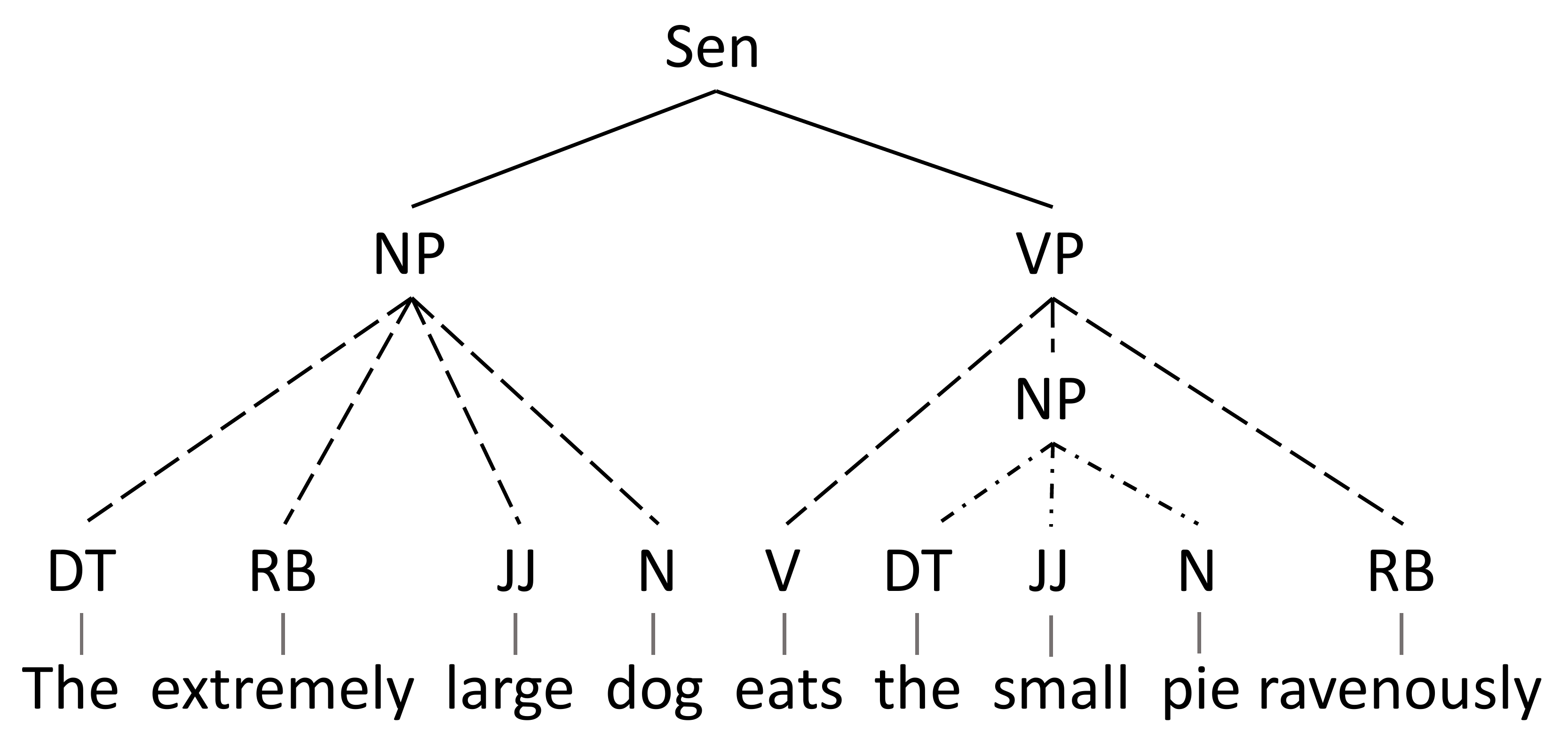}
\caption{An illustration of generating a syntax tree for a source sentence. In the first iteration, ``Sen'' consists of (``NP'', ``VP'') as the solid lines. In the second iteration, ``NP'' consists of (``DT'', ``RB'', ``JJ'', ``N'') and ``VP'' consists of (``V'', ``NP'', ``RB'') as the dash lines. In the third iteration, ``NP'' consists of (``DT'', ``JJ'', ``N'') as the dot-and-dash lines.}
\label{fig:constree}
\end{figure}

\begin{figure*}[t]
\centering
\includegraphics[width=0.8\textwidth]{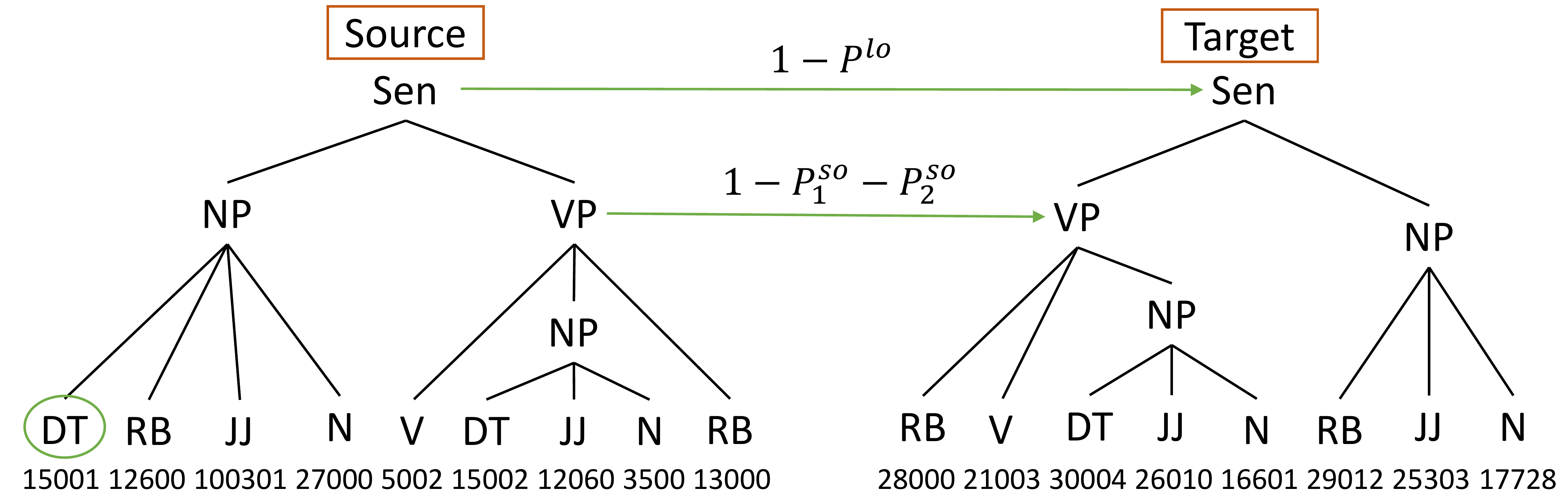}
\caption{An illustration of ``translation'', where the constituent order of ``Sen'' is changed to ``VP NP'' with probability $1-P^{lo}$, the constituent order of ``VP'' is changed to ``RB V NP'' with probability $1-P_1^{so}-P_2^{so}$, and the circled ``DT'' is removed with probability $P^{op}$. Meanwhile, the numbers in the source sentence are replaced with the ones in the target sentence based on mappings.}
\label{fig:changetree}
\end{figure*}

We first employ phrase structure rules~\cite{chomsky1959certain} to synthesize the source sentences, where the rules are based on the syntax of languages. Considering that translation can be decomposed to word reordering and word translation~\cite{bangalore2001finite,sudoh2011post}, we then ``translate'' the synthesized source sentences to synthesized target sentences in two steps: 1) word reordering by changing its syntax tree; 2) and word translation by substituting the source words into target words.

\paragraph{Source Sentence Synthesis.} We first generate the syntax tree of the source sentence. Specifically, we use the notations of the constituents in syntax tree according to the Penn Treebank syntactic and part of speech (POS) tag sets\footnote{``Sen'':sentence; ``NP'': noun phrase; ``VP'': verb phrase; ``DT'': determiner; ``JJ'': adjective; ``RB'': adverb; ``N'': noun; ``V'': verb.}~\cite{marcus1993building}, and generate the syntax tree of a source sentence as following~\cite{rosenbaum1967phrase}:
\begin{itemize}
    \item Sen $\to$ NP VP,
    \item NP $\to$ $\left(\mbox{DT}\right)$ $\left(\mbox{RB}\right)^\ast$ $\left(\mbox{JJ}\right)^\ast$ N,
    \item VP $\to$ V $\left(\mbox{NP}\right)$ $\left(\mbox{RB}\right)^\ast$,
\end{itemize}
where the constituent on the left side of the arrow consists of the constituents on the right side in sequence, ``($\cdot$)'' means that the constituent is optional, and ``($\cdot$)*'' denotes that the constituent is not only optional but can also be repetitive. For each sentence, we start with a single constituent Sen and iteratively decompose ``Sen'', ``NP'', and ``VP'' according to the rules until all the constituents are decomposed to ``DT'', ``JJ'', ``RB'', ``V'', and ``N''. An illustration of generating a syntax tree is depicted in Fig.~\ref{fig:constree}. To synthesize the source sentence according to the syntax tree, we use numbers as the words in the synthesized source sentences and use different ranges of numbers to represent words with different POS, where the details of the ranges are provided in Appendix \ref{app:A}. Then, a number is randomly sampled in the corresponding range for each word in the syntax tree.

\paragraph{Word Reordering.} To introduce syntactic multi-modality, we consider multiple possible rules for ``Sen'', ``NP'', and ``VP'' in the target sentences. Dependency distance is defined as the linear distance between two words with syntactical relationship~\cite{liu2017dependency}, which can be used as a guide to select typical rules to introduce long- and short-range word order diversity. Specifically, we consider three options: 1) The word order of ``Sen'' is with probability $P^{lo}$ to be the same with the source sentence (i.e., NP VP) and with probability $\left(1-P^{lo}\right)$ to swap the ``NP'' and ``VP'' (i.e., VP NP), which has long dependency distance and represents for the long-range word order; 2) For the word order in ``VP'', it is considered to be the same with the source sentence with probability $P^{so}_{1}$, place ``RB'' between ``V'' and ``NP'' with probability $P^{so}_2$, and place ``RB'' before ``V'' with probability $\left(1-P^{so}_{1}-P^{so}_2\right)$, which has short dependency distance and represents for the short-range word order; 3) To introduce the syntactic multi-modality of optional words, we change the existence of ``DT'' in each ``NP'' of the source sentence with probability $P^{op}$ (i.e, remove ``DT'' if it exists in the source sentence and add ``DT'' if it does not exist in the source sentence).

\paragraph{Word Translation.} Same as in the source sentences, we use different range of numbers to represent words with different POS in target sentences. To do the word translation, we first randomly build mappings between the source and target words with different POS respectively. Since we focus on studying the syntactic multi-modality, we consider each source word is mapped to a single target word to eliminate the lexical multi-modality. Then, we replace the words in the source sentence based on the mappings to generate the target sentence. An illustration of ``translation'' is shown in Fig.~\ref{fig:changetree}.

\subsection{Experiments and Analyses}
\begin{table}[]
\begin{tabular}{c|c|c}
\toprule
Probability         & Default      & Effect            \\ 
\midrule
$P^{lo}$         & 1      & long-range word order              \\ 
\midrule
$P^{so}_1$       & 1      & short-range word order              \\ 
\midrule
$P^{so}_2$       & 0      & short-range word order              \\ 
\midrule
$P^{op}$         & 0      & optional words            \\ 
\bottomrule
\end{tabular}
\caption{Default values of the probabilities to adjust the syntactic multi-modality.}
\label{tab:probability}
\end{table}

We conduct experiments to compare existing loss functions on different kinds of syntactic multi-modality on the synthesized datasets, by changing the probabilities (i.e., $P^{op}$, $P^{so}_{1}$, $P^{so}_{2}$, and $P^{lo}$) as listed in Table~\ref{tab:probability}. In the following, we first provide the experimental settings, then show the results on the long-range and short-range syntactic multi-modalities, and finally conclude the key findings.

\paragraph{Experimental Settings.} We consider two separate vocabularies for the source and target sentences, each containing $15$K words. $0.3$M, $5$K, and $5$K synthesized sentence pairs are generated as training, validation, and test sets respectively. We use the same hyper-parameters in the transformer-base model~\cite{vaswani2017attention}, which is commonly used in the NAT models~\cite{gu2018non,du2021order,saharia2020non}. All 
settings
are trained on $4$ Nvidia V100 GPUs with $16$k tokens in a batch. For the model with OAXE loss, we train the first $50$K updates with XE loss and the next $50$K updates with OAXE loss~\cite{du2021order}. For the other losses, we train the model for $100$K updates. The length of the decoder input is set as twice the length of the source sequence for CTC loss~\cite{saharia2020non}, while the golden target length is used for OAXE, AXE, and XE. To evaluate the accuracy of the predicted sequence, we first calculate the longest common sub-sequence between the predicted and the golden sequences, and then the accuracy is defined as the ratio between the length of the longest common sub-sequence and the golden sequence. The accuracy on the test set is calculated as the average accuracy among all the predicted sentences.

\begin{figure}[t]
\centering

\begin{subfigure}[t]{0.48\textwidth}
    \includegraphics[width=\textwidth]{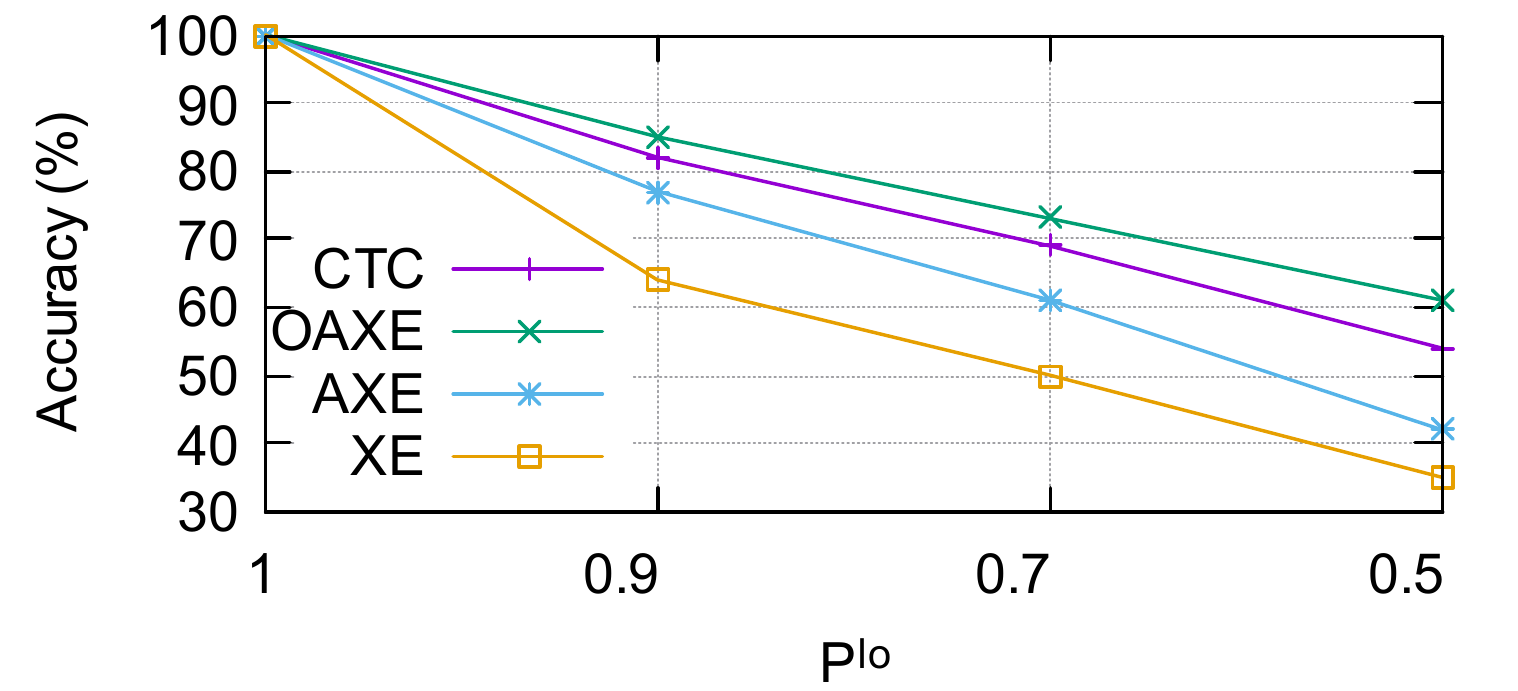}
    \caption{Effect of long-range word order.}
    \label{fig:lo}
\end{subfigure}

\begin{subfigure}[t]{0.48\textwidth}
    \includegraphics[width=\textwidth]{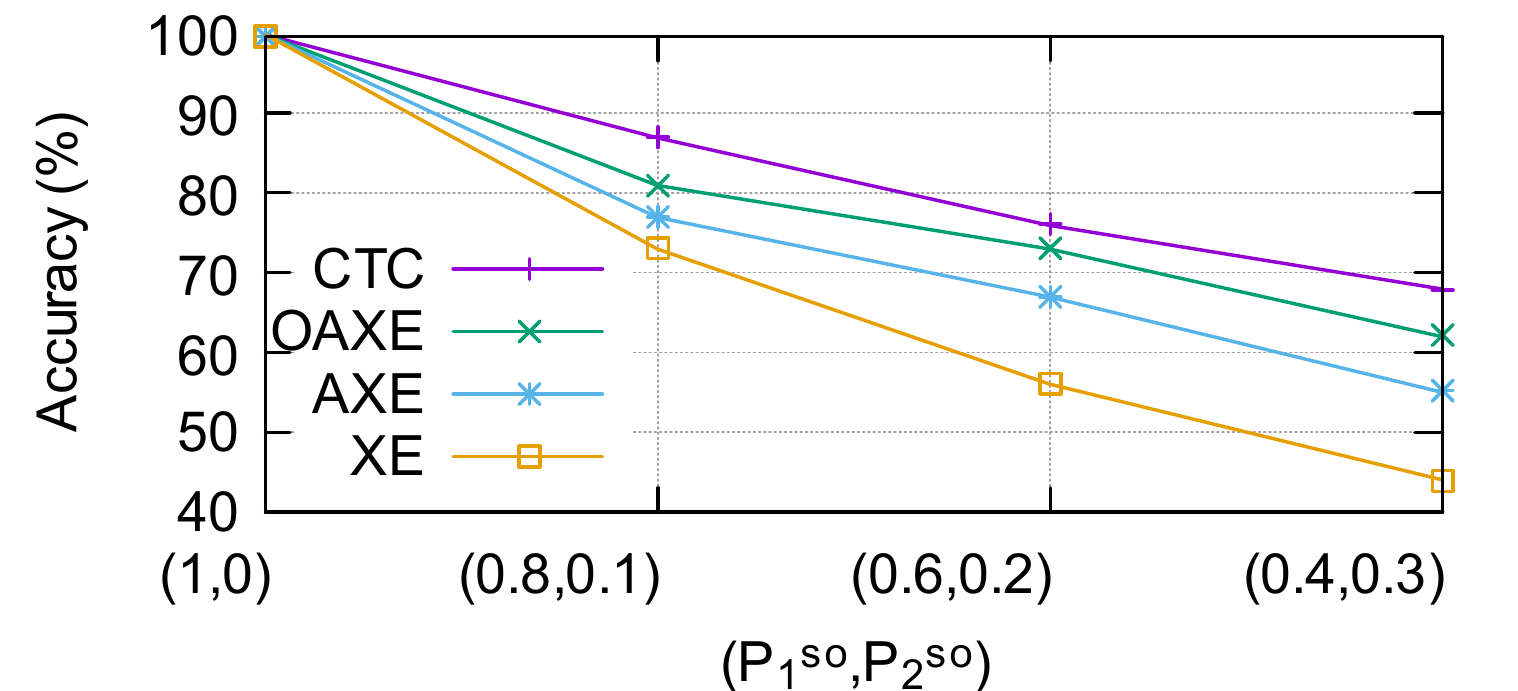}
    \caption{Effect of short-range word order.}
    \label{fig:so}
\end{subfigure}

\begin{subfigure}[t]{0.48\textwidth}
    \includegraphics[width=\textwidth]{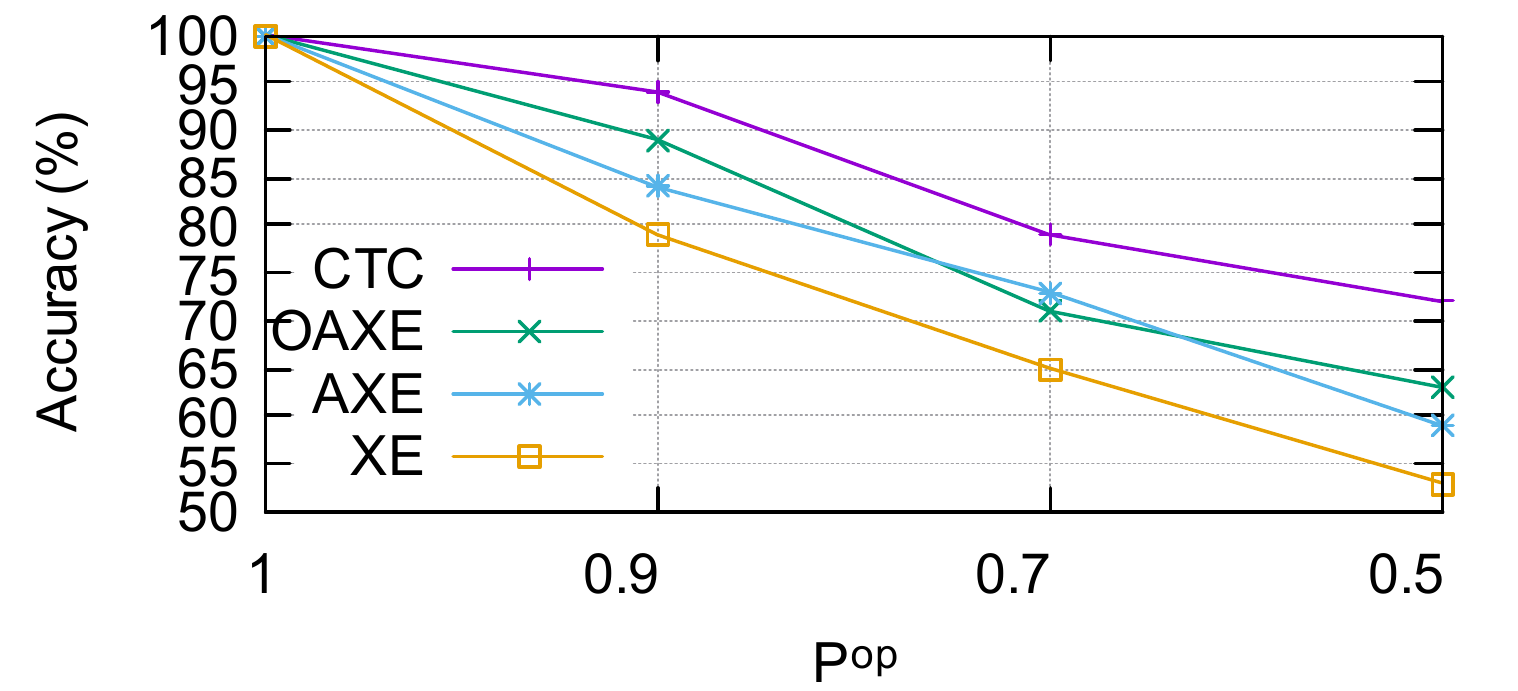}
    \caption{Effect of optional words.}
    \label{fig:op}
\end{subfigure}

\caption{The accuracy of different loss functions on synthesized datasets.}
\vspace{-10pt}
\label{fig:toy}
\end{figure}

\paragraph{Long-Range Syntactic Multi-modality.} To consider the effect of long-range diversity, we change the corresponding probability $P^{lo}$, while keeping the others unchanged to eliminate the short-range syntactic multi-modality. It is observed in Fig.~\ref{fig:lo} that CTC loss always performs better than AXE, and OAXE is the best with different degree of long-range multi-modality.


\paragraph{Short-Range Syntactic Multi-modality.} Similarly, we only change the probabilities $P^{so}_{1}$ and $P^{so}_{2}$ to adjust the degree of short-range word order diversity. The results are shown in Fig.~\ref{fig:so}, where OAXE loss performs better than AXE loss, and CTC loss outperforms all the other losses with varied degree of short-range word order diversity. In order to study the effect of optional words, we vary the probability $P^{op}$ to change the existence of ``DT''. As shown in Fig.~\ref{fig:op}, OAXE loss is slightly better than AXE loss, and CTC loss performs the best, indicating that CTC loss is superior in the syntactic multi-modality problem caused by optional words.



\paragraph{Analyses and Discussions.} Based on the results in Fig.~\ref{fig:toy}, we can get the following observations:
\begin{itemize}[leftmargin=*]
    \item OAXE loss is superior in handling the long-range syntactic multi-modality (i.e., long-range word order). OAXE loss is order-agnostic, which is able to provide fully positive feedback to the word in different positions compared to the ground truth sequence. Accordingly, OAXE is suitable for datasets with long-range word order diversity. Though it can deal with high diversity of word order, it may also incur wrong predictions on word order, which may be why OAXE is not suitable when the diversity only exists in short-range.
    \item CTC loss is the best choice for dealing with short-range syntactic multi-modality (i.e., short-range word order and optional words). CTC loss is generally considered to handle monotonic matching, which seems not effective in handling the multi-modality caused by word order~\cite{saharia2020non}. However, it is observed in Fig.~\ref{fig:lo} and~\ref{fig:so} that CTC loss outperforms AXE and XE when dealing with long-range word order diversity and performs the best on the multi-modality caused by short-range word order. Since CTC considers all the monotonic alignments, it can partially provide positive feedback to the words with different order through multiple monotonic alignments. As shown in Fig.~\ref{fig:CTC}, all the words can be considered in the three alignments. 
\end{itemize}
Considering that AXE loss does not show superiority on any type of the syntactic multi-modality, we will only focus on CTC and OAXE losses in the following.

\section{Analyses on Real Datasets}
\label{sec:real}
Though carefully designed, the synthesized sentence pairs consisting of numbers are still different from the real sentence pairs.
Therefore, in this section, we validate the findings in Section~\ref{sec:toy} based on real datasets. Considering that different types of syntactic multi-modality are highly coupled in the real corpus, we 
conduct experiments on carefully selected sub-datasets
from a corpus, to approximately
decompose the syntactic multi-modality. In the following, we first show the details of the approach to decompose the syntactic multi-modality, and then provide the analytical results based on the real datasets.





\paragraph{Analytical Approach.} In order to decompose the long-range and short-range types of syntactic multi-modality, we select sentences that only contain subject and verb phrases from a corpus, and divide them into two sub-datasets according to the relative order of subject and verb (i.e., subject first that denoted as ``SV'', or verb first that denoted as ``VS''). Meanwhile, we only consider the declarative sentence pairs in the corpus to eliminate the word order difference caused by mood. Following this method, the long-range multi-modality is eliminated in each sub-dataset (i.e., ``SV'' and ``VS''), which can be used to evaluate the effect of short-range multi-modality. To analyse the long-range multi-modality, we can adjust the degree of long-range word order diversity by sampling data from the two sub-datasets with varied ratios, while roughly keeping the degree of short-range word order diversity unchanged. Specifically, considering that Russian is flexible on word order~\cite{kallestinova2007aspects} and it is feasible to select sentences on both the ``SV'' and ``VS'' order, we use an English-Russian (EN-RU) corpus from Yandex\footnote{https://translate.yandex.ru/corpus} 
that contains
$1$M EN-RU sentence pairs, from which we get $0.2$M and $0.1$M sentence pairs data with ``SV'' order and ``VS'' order respectively. To select the sentence pairs with different word orders, we use spaCy\cite{Honnibal_spaCy_Industrial-strength_Natural_2020} to parse the dependency of Russian sentences. For the models with CTC loss, we train for $300$K updates. For the models with OAXE loss, we train with XE loss for $100$K updates and then train with OAXE loss for $200$K updates.



\begin{table}[t]
  \centering
  \caption{BLEU scores of models with CTC and OAXE losses, where the models are evaluated on the WMT'19 EN-RU test set. The percentage of sentences with ``RB V'' among the sentences with both ``RB V'' and ``V RB'' orders are shown in column ``RB V''. The percentage of sentences with ``JJ N'' among the sentences with both ``JJ N'' and ``N JJ'' orders are shown in column ``JJ N''.}
  \label{tab:fulldata}
  \vspace{-10pt}
\begin{tabular}{c|cc|cc}
\toprule 
``SV'':``VS'' & CTC & OAXE & ``RB V'' & ``JJ N'' \\
\midrule 
$100\%$ : $0\%$  & 17.7 & 16.5 & 68\% & 84\% \\
$75\%$ : $25\%$  & 17.2 & 16.9 & 63\% & 82\% \\
$50\%$ : $50\%$  & 16.8 & 17.3 & 70\% & 79\% \\
\bottomrule
\end{tabular}
\vspace{-10pt}
\end{table}

\paragraph{Analytical Results.} We keep the total number of sentence pairs in the training set as $0.2$M (i.e., the number of Russian sentences in the ``VS'' sub-dataset), and change the ratio of sentence pairs sampled from two sub-datasets (i.e., ``SV'' and ``VS''). The results are shown in Table~\ref{tab:fulldata}, where the training parameters are the same as that used in Section~\ref{sec:toy}. It is observed that CTC loss outperforms OAXE loss when all data samples are from the ``SV'' sub-dataset, which indicates that CTC loss performs better on short-range syntactic multi-modality problem. When the ratio of the data sizes on the two sub-datasets is changed to $75\%$ : $25\%$, the gap between the performance of CTC and OAXE losses diminished, while CTC loss still performs slightly better than OAXE loss. When the ratio changed to $50\%$ : $50\%$, model with OAXE loss becomes better than that with CTC loss. In summary, OAXE loss is better at handling long-range syntactic multi-modality while CTC loss is better on short-range syntactic multi-modality, which validates the key observations we obtained on the synthesized datasets in Section~\ref{sec:toy}.

In order to demonstrate whether we have decomposed the long- and short-range syntactic multi-modalities, we verify whether the degree of short-range multi-modality remains almost the same when varying the degree of long-range multi-modality. We evaluate the short-range syntactic diversity based on the relative order between: 1) adverb and verb (``RB V''); 2) adjective and noun (``JJ N''). As shown in Table~\ref{tab:fulldata}, when the ratio of the data sizes on the two sub-datasets varied from $100\%$ : $0\%$ to $50\%$ : $50\%$ (i.e., the ratio between ``SV'' and ``VS'' changes),  the relative order on ``RB V'' and ``RB V'' (which can represent the degree of short-range word order diversity) does not vary much. These analyses verify the rationality of our analytical approach in this section.

\section{Better Solving the Syntactic Multi-Modality Problem}
\label{sec:better_solve}

As shown in previous sections, the CTC and the OAXE loss functions are good at dealing with short- and long-range syntactic multi-modalities respectively. While in real-world corpora, different types of multi-modalities usually occur together and vary in different languages. Accordingly, it may be better to use different loss functions for different languages. In this section, we first introduce a new loss function named Combined CTC and OAXE (CoCO), which takes advantage of both CTC and OAXE to better handle the long-range and short-range syntactic multi-modalities simultaneously, and then provide a guideline on how to choose the appropriate loss function for different scenarios.


\subsection{CoCO Loss}
To obtain a general loss that performs well at both types of multi-modalities, it is natural to combine the two loss functions studied above. However, the output length is mismatched between the models using CTC and OAXE, where the output length is required to be longer than the target sequence with CTC loss, and is required to be the same as the target sequence with OAXE loss. To solve this length mismatch problem, we consider using the same output length as in CTC loss, and
modify the OAXE loss to make it suitable on this output length by allowing consecutive tokens in the output to be aligned with the same token in the reference sequence. The details of the modified OAXE loss are provided in Appendix \ref{app:B}. Then, the proposed CoCO loss is defined as a linear combination of the CTC and modified OAXE losses as:
\begin{equation}
    \mathcal{L}_{CoCO} = \lambda \mathcal{L}_{CTC} + ( 1- \lambda) \mathcal{L}_{M-OAXE},
\end{equation}
where $\mathcal{L}_{M-OAXE}$ denotes the modified OAXE loss and $\lambda$ is a hyper-parameter that balances the two losses.

\subsection{Choosing Appropriate Loss Function}
\begin{figure}[t]
\centering
\includegraphics[width=0.48\textwidth]{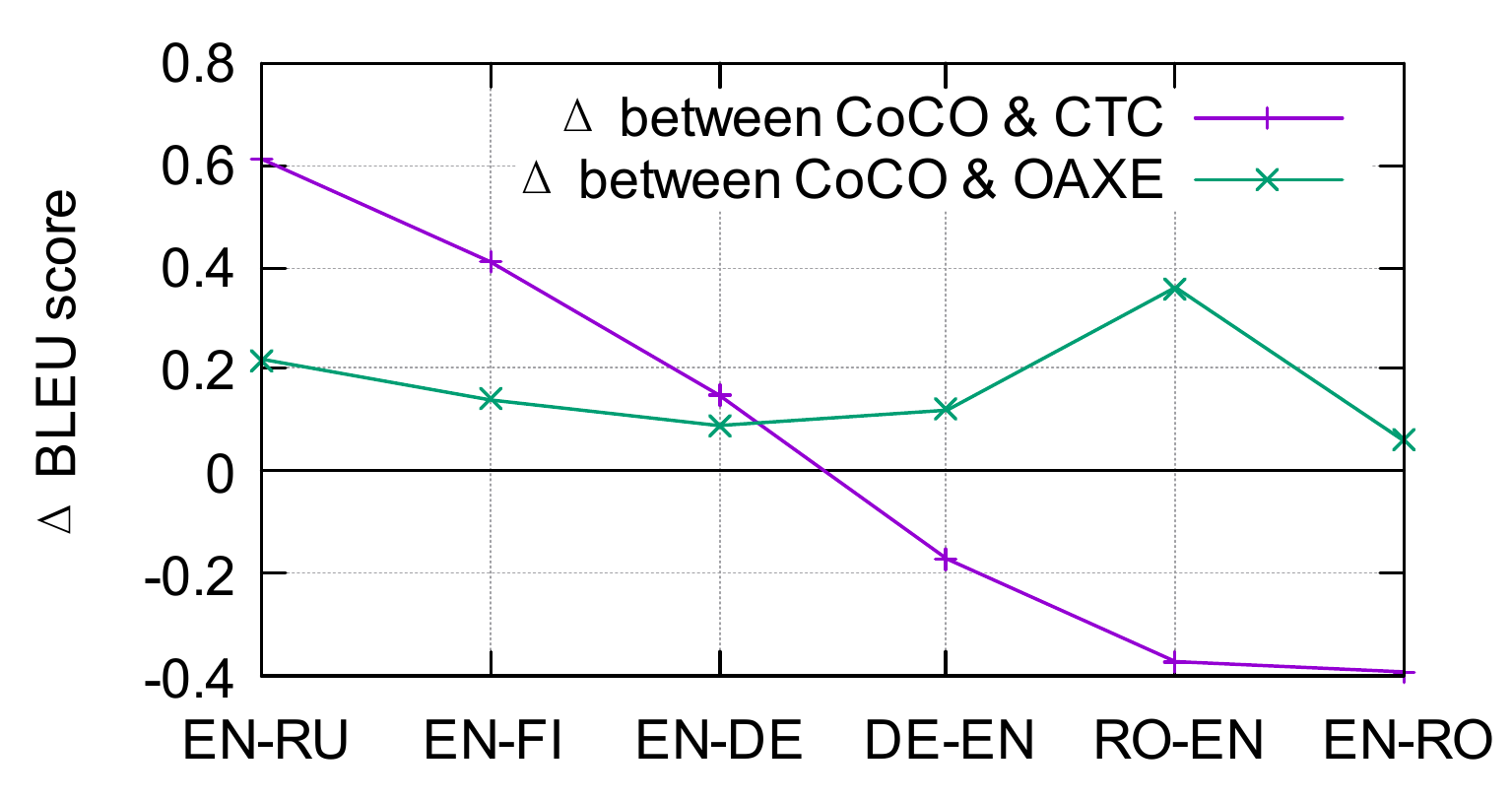}
\vspace{-10pt}
\caption{Comparing different losses on different language pairs.}
\vspace{-10pt}
\label{fig:languages}
\end{figure}

The degree of different types of multi-modalities varies among different languages. In order to find the insight to choose the appropriate loss function for different languages, we conduct experiments on several languages including Russian (RU), Finnish (FI), German (DE), Romanian (RO), and English (EN). These languages have different requirements on the positions of subject (S), verb (V), and object (O), which is one major influence factor on the large-range syntactic multi-modality. Specifically, the order in RU and FI is quite flexible, where all the 6 possible orders of ``S'', ``V'', and ``O'' are valid. In DE, the verb is required to be placed on the second position, which is called verb-second word order. Meanwhile, in RO and EN, the order is restricted to ``SVO''.


\begin{table*}[tb]
\centering
  \caption{BLEU scores of NAT models.}
  \vspace{-10pt}
  \label{tab:startoftheart}
\begin{tabular}{l|ccccc|c}
\toprule 
\multirow{2}{*}{\textbf{Model}} & \multicolumn{2}{c}{WMT14} & \multicolumn{1}{c}{WMT16} & \multicolumn{1}{c}{WMT14} & \multicolumn{1}{c}{WMT17} &  \\
                  &   EN-DE  &   DE-EN  &    EN-RO  & EN-RU  &  EN-FI  & Speedup \\
\midrule
\textbf{Autoregressive}&             &           &             &           \\
\quad Transformer&     27.48     & 31.39  &      33.70         &   27.2   & 28.12  & 1.0 $\times$  \\
\midrule
\textbf{Non-Autoregressive}&           &           &           &           \\
\quad  Vanilla NAT\small{~\cite{gu2018non}}&   17.69 & 21.47  & 27.29      & --  & -- & 15.0 $\times$   \\
\quad  BoN\small{~\cite{shao2020minimizing}}&   20.90 &  24.60 & 28.30      & --  & --  & 10.0 $\times$ \\
\quad  AXE\small{~\cite{ghazvininejad2020aligned}}&   23.53 & 27.90 & 30.75       & --  & --  & 15.3 $\times$ \\
\quad  Imputer\small{~\cite{saharia2020non}}&   25.80   & 28.40 & 32.30      & --  & --   & 18.6 $\times$  \\
\quad  OAXE (CMLM)\small{~\cite{qian2020glancing}}& 26.10  & 30.20 & 32.40       & --  &  -- &  15.6 $\times$ \\
\quad  GLAT\small{~\cite{qian2020glancing}}&   26.39  & 29.84 & 32.79       &  -- & --  & 14.6 $\times$ \\
\quad  CTC (VAE)\small{~\cite{gu2020fully}}&   27.49 & 30.46 & 33.79       & --  & --  & 16.5 $\times$ \\
\quad  CTC (GLAT)\small{~\cite{gu2020fully}}&   27.20  & 31.39 & 33.71         & -- & --  & 16.8 $\times$  \\
\quad  CTC (DSLP)\small{~\cite{huang2021non}}&   27.02  & 31.61 & 34.17        &   21.38  &  22.83  &  14.8 $\times$ \\
\quad  CoCO (DSLP) &      27.41     &   31.37         &      34.32     &      21.82  &    23.25 &    14.2 $\times$ \\
\bottomrule
\end{tabular}
\vspace{-10pt}
\end{table*}

We evaluate the accuracy of different loss functions (i.e., CTC, OAXE, and CoCO) on WMT'14 EN-RU, WMT'17 EN-FI, WMT'14 EN-DE, and WMT'16 EN-RO datasets with 1.5M, 2M, 4M, and 610K sentence pairs, respectively. The $\lambda$ in COCO loss is set as $0.1$ so that $\lambda \mathcal{L}_{CTC}$ and $( 1- \lambda)\mathcal{L}_{M-OAXE}$ are in the same order of magnitude. Following \citet{du2021order}, for the models with OAXE and CoCO loss, we first train with XE or CTC loss for $100$K updates and then train with OAXE or CoCO loss for $200$K updates, respectively. For CTC loss, we train for $300$K updates. For decoding, we follow \citet{gu2020fully,huang2021non} to use beam search with language model scoring\footnote{https://github.com/kpu/kenlm} for CTC and CoCO. The other training settings are the same as used in Section~\ref{sec:toy}. 
We report the tokenized BLEU score to keep consistent with previous works.
We show the difference values of BLEU score in Fig.~\ref{fig:languages} and provide the corresponding absolute BLEU scores in Appendix \ref{app:C}. According to Fig.~\ref{fig:languages}, we have several observations: 1) The proposed CoCO loss consistently improves the translation accuracy on all the language pairs compared to OAXE loss; 2) The CoCO loss outperforms CTC loss when the target language is with flexible word order or verb-second word order (i.e., EN-RU, EN-FI, and EN-DE); 3) CTC loss performs the best when the target language is ``SVO'' language (i.e., DE-EN, RO-EN, and EN-RO).

We would also like to evaluate the CoCO loss based on SOTA NAT models. Though the proposed CoCO loss can be used in both iterative and non-iterative models, we only show the results on non-iterative models in this paper and leave the iterative models as future work. We use CoCO loss on a recently proposed Deeply Supervised, Layer-wise Prediction-aware (DSLP) transformer~\cite{huang2021non}, which achieves competitive results. The details of how CoCO loss is applied on DSLP are provided in Appendix \ref{app:D}. The results are shown in Table~\ref{tab:startoftheart}. Compared to DSLP with CTC loss~\cite{huang2021non}, DSLP with CoCO loss consistently improves the BLEU scores on three language pairs, including EN-RU, EN-FI, and EN-DE. On the contrary, DSLP with CTC loss is better or comparable to DSLP with CoCO loss when the target language is restricted to the ``SVO'' word order, including EN-RO and DE-EN.


According to the experiments on language pairs with different kinds of requirements on word order, we suggest to: 1) use CoCO loss when the word order of the target language is relatively flexible ( e.g., RU and FI, where word order on ``S'' ``V'' ``O'' is free, and DE, where the verb is required to be placed on the second position); 2) use CTC loss when the target language is with relatively strict word order (e.g., RO and EN, which are ``SVO'' languages).

\section{Conclusion}
In this paper, we conduct a systematic study on the syntactic multi-modality problem in non-autoregressive machine translation. We first categorize this problem into long-range and short-range types and study the effectiveness of different loss functions on each type. Considering the different types are usually entangled in real-world datasets, we design and construct synthesized datasets to control the degree of one type of multi-modality without changing another for analyses. We find that CTC loss is good at short-range syntactic multi-modality while OAXE loss is better at the long-range one. These findings are further verified on real-world datasets with our designed analytical approach. Based on these analyses, we propose a CoCO loss that can better handle the complicated syntactic multi-modality in real-world datasets, and a practical guide to use different loss functions for different kinds of syntactic multi-modality: CoCO loss is preferred when the word order of target language is relatively flexible while CTC loss is preferred when target language is with strict word order. Our study in this paper can facilitate better understanding of the multi-modality problem and provide insights to better solve this problem in non-autoregressive translation. Besides, there still remain some open problems that need future investigation. For example, we generally consider long-range and short-range types for syntactic multi-modality, while there may be more fine-gained categorizations on the syntactic multi-modality due to the complex syntax of natural language.


\bibliography{ref}

\newpage
\section*{Appendix}
\appendix
\section{Number Ranges to Synthesis the Source and Target Sentences}\label{app:A}
We use [1, 5000], [5001, 10000], [10001, 12500], [12501, 15000], and {15001, 15002, 15003} to represent ``N'' ``V'' ``JJ'' ``RB'' ``DT'' in the source sentences, and [15004, 20003], [20004, 25003], [25004, 27503], [27504, 30003], and {30004, 30005, 30006} to represent ``N'' ``V'' ``JJ'' ``RB'' ``DT'' in the target sentences.

\section{Modified OAXE Loss}\label{app:B}
\begin{figure}[h]
\centering
\includegraphics[width=0.48\textwidth]{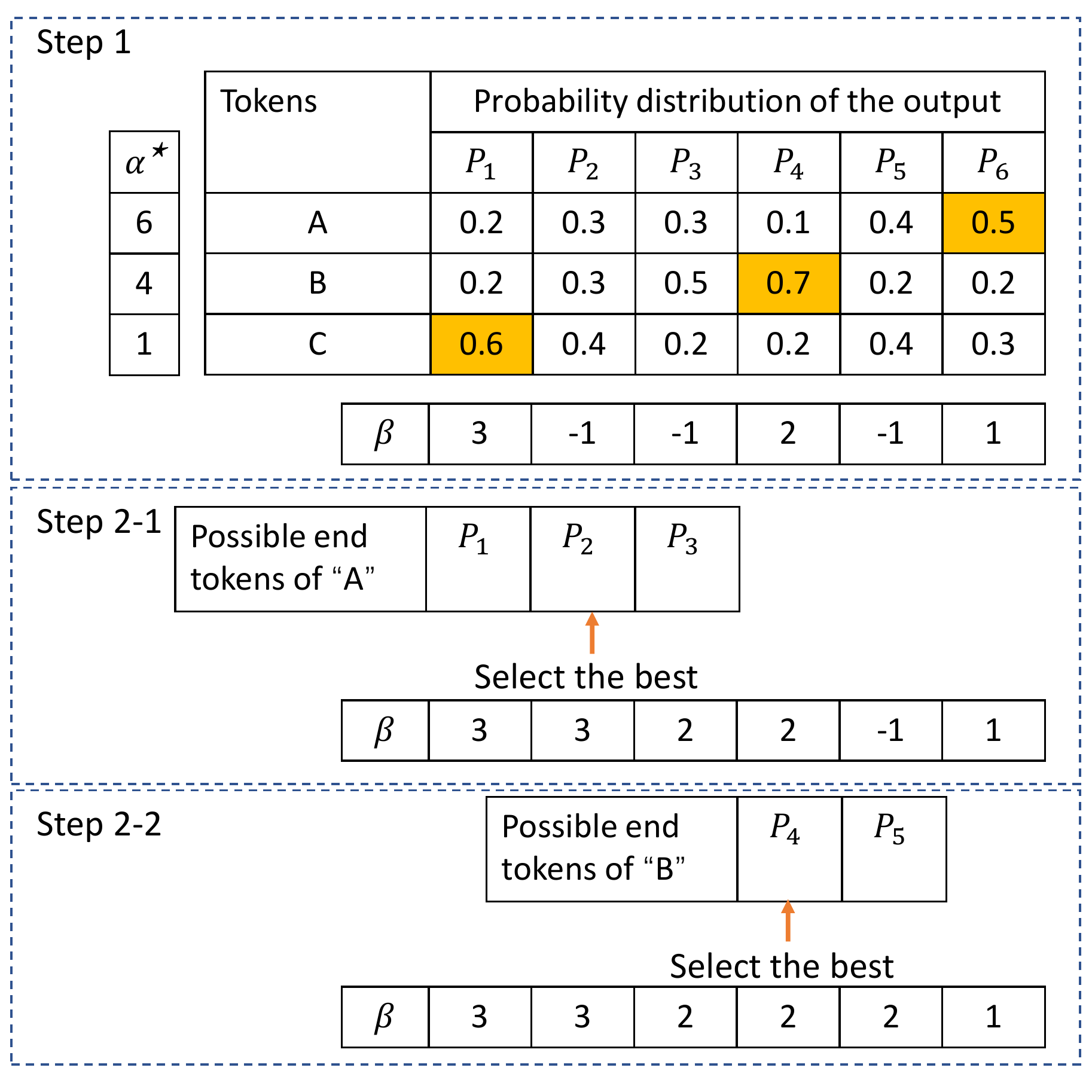}
\caption{An illustration of the modified OAXE loss.}
\label{fig:M-OAXE}
\end{figure}
\begin{table*}[t]
  \centering
  \caption{BLEU scores of models with different losses on different language pairs.} 
  \label{tab:languages}
\begin{tabular}{l|cccccccc}
\toprule 
Loss & EN-RU & EN-FI & EN-DE & DE-EN & RO-EN & EN-RO \\
\midrule 
CTC & 20.84 & 22.86 & 26.10 & 30.36 & 33.68 & 33.06 \\
OAXE & 21.23 & 23.13 & 26.16 & 30.07 & 33.25 & 32.31 \\
CoCO & 21.45 & 23.27 & 26.25 & 30.19 & 33.31 & 32.67\\
\bottomrule
\end{tabular}
\end{table*}
Specifically, we consider one training pair ($X$,$Y$), where there are $n$ tokens in the ground truth sequence, denoted as $Y = (y_1,y_2,\dots,y_n)$. The corresponding output sequence has $m$ tokens with probability distributions $P_1,P_2,\dots,P_m$, where $m>n$. According to OAXE, we first get the alignment between the ground truth sequence and the output sequence that minimizes the cross entropy loss based on maximum bipartite matching algorithm~\cite{kuhn1955hungarian}:
\begin{equation}\label{eq:align}
    \alpha^{\star} = \mathop{\arg\min}\limits_{\alpha \in \gamma (\alpha) } \left( -\sum \limits_{i}\log P_{\alpha(i)}(y_i|X,\theta) \right),
\end{equation}
where $\alpha$ denotes the alignment from the ground truth sequence to the output sequence, $\gamma (\alpha)$ is the set of all possible alignments, and $y_i$ is aligned with the $\alpha(i)$-th token of the output. We consider each output token can only be aligned to one ground truth token (i.e., $\alpha(i) \neq \alpha(j)$ if $i \neq j$). Then, we can get the alignment from the output sequence to the ground truth sequence, based on $\alpha^\star$:
\begin{equation}
    \beta(k) = \left\{
    \begin{aligned}
        i & \text{\quad if } \alpha^\star(i) = k, \\
        -1 & \text{\quad if } \forall i \in [1,n], \alpha^\star(i) \neq k,
    \end{aligned}
    \right.
\end{equation}
where the $k$-th token of the output is aligned to $y_{\beta(k)}$ and $\beta(k) = -1$ denotes the token has not been aligned. We provide an illustration 
as the ``step 1'' in Fig.~\ref{fig:M-OAXE}, where we consider 3 tokens in the target sequence and 6 tokens in the output and the best alignment is ``A''-``$P_6$'', ``B''-``$P_4$'', and ``C''-``$P_1$''. Since consecutive repetitive tokens are merged when decoding with CTC loss, we consider that consecutive tokens in the output can be aligned to the same ground truth token. Accordingly, we enumerate the end of each ground truth token in the output sequence respectively, and select the one that minimize the cross entropy loss. For example, given $\beta(k_1) = i$, $\beta(k_2) = j$ and $\beta(k) = -1$ when $k_1 \le k \le k_2$, we select $k^\star$ according to:
\begin{equation}
    \begin{aligned}
    k^\star =  \mathop{\arg\min}\limits_{k_1 \le k' < k_2} \bigg(& - \sum \limits_{k_1 \le k \le k'} \log P_k(y_i|X,\theta) \\
    &- \sum \limits_{k' < k \le k_2} \log P_k(y_j|X,\theta) \bigg),
    \end{aligned}
\end{equation}
and align the $(k_1,k^\star]$-th output token to $i$ and the $(k^\star,k_2)$-th output token to $j$ as:
\begin{equation}
    \beta(k) = \left\{
    \begin{aligned}
        i & \text{\quad if } k \in (k_1,k^\star] \\
        j & \text{\quad if } k \in (k^\star,k_2).
    \end{aligned}
    \right.
\end{equation}
As the illustration in Fig.~\ref{fig:M-OAXE}, we enumerate all the possible end tokens of 'A' and 'B' to find the best one.
Then, we get the modified OAXE loss as:
\begin{equation}
    \mathcal{L}_{M-OAXE} = -\sum \limits_{1 \le k \le m}\log P_k \left( y_{\beta(k)}|X,\theta \right).
\end{equation}


\section{BLEU Scores of Different Losses on Different Language Pairs.}\label{app:C}
The BLEU scores of models with CTC, OAXE and CoCO loss on different languages pairs are shown in Table~\ref{tab:languages}.

\section{Use CoCO Loss in DSLP}\label{app:D}

Partially feeding ground truth tokens to the decoder during training shows promising performance in NAT~\cite{ghazvininejad2019mask,saharia2020non,qian2020glancing,huang2021non}. For the models training with golden length of the ground truth sentence using XE loss, the ground truth token embedding is placed to the position of the corresponding input~\cite{qian2020glancing}. When using CTC loss, the inputs of the decoder are always longer than the ground truth sentences, where \citet{gu2020fully} proposes to use the best monotonic alignment between the ground truth and output sequences, and provides the ground truth to the corresponding input position of the decoder. With the proposed CoCO loss, we use the best alignment which is not required to be monotonous. In addition, DSLP requires deep supervision on each layer of the decoder. We find that only replacing CTC loss with CoCO loss on the first layer is better than using CoCO loss on all layers. Accordingly, when using CoCO loss in DSLP transformer, we use CoCO loss in the first layer and CTC loss for all the other layers in the DSLP transformer.

\end{document}